\documentclass[final]{cvpr}

\usepackage{epsfig}
\usepackage{url}
\usepackage{cite}
\usepackage{balance}
\usepackage{graphicx}
\usepackage{amsfonts}
\usepackage{fancyhdr}
\usepackage{comment}
\usepackage{times}
\usepackage{amsmath}
\usepackage{changepage}
\usepackage{amssymb}
\usepackage{enumerate}
\usepackage{algorithmic}
\usepackage{bm}
\usepackage{calligra}
\usepackage{multirow}
\usepackage{tabularx}
\usepackage{booktabs}
\usepackage{mathtools}
\usepackage{arydshln}
\usepackage{latexsym} 
\usepackage{amsmath}
\usepackage{amssymb}
\usepackage{color}
\usepackage[ruled,vlined]{algorithm2e}
\usepackage{subfigure}
\usepackage{ulem}
\usepackage{float}
\usepackage{tabstackengine}
\usepackage{siunitx}
\usepackage{colortbl}
\usepackage{hhline}
\usepackage{textcomp}

\newcommand{\rt}{\textcolor[rgb]{1,0,0}}

\usepackage[pagebackref=true,breaklinks=true,colorlinks,bookmarks=false]{hyperref}

\def\cvprPaperID{18} 
\def\confYear{CVPR Workshop 2021}

\begin{document}

\title{Comparing Representations in Tracking for Event Camera-based SLAM}
\author{Jianhao Jiao$^{\dag}$\ \ \ \ \ Huaiyang Huang$^{\dag}$\ \ \ \ \ Liang Li$^{\ddag}$\ \ \ \ \ Zhijian He$^{\dag}$\ \ \ \ \ Yilong Zhu$^{\dag}$\ \ \ \ \ Ming Liu$^{\dag}$\\
${\dag}$ The Hong Kong University of Science and Technology\thanks{This work was supported by Collaborative Research Fund by Research Grants Council Hong Kong, 
under Project No. C4063-18G, Department of Science and Technology of Guangdong Province Fund, under Project No. GDST20EG54 and Zhongshan Municipal Science and Technology Bureau Fund, under project ZSST21EG06, awarded to Prof. Ming Liu. Email: jjiao@connect.ust.hk}
\\
${\ddag}$ The University of Hong Kong
}

\maketitle

\begin{abstract}
This paper investigates two typical image-type representations for event camera-based tracking: time surface (TS) and event map (EM).
Based on the original TS-based tracker, we make use of these two representations' complementary strengths to develop an enhanced version.
The proposed tracker consists of a general strategy to evaluate the optimization problem's degeneracy online and then switch proper representations.
Both TS and EM are motion- and scene-dependent, and thus it is important to figure out their limitations in tracking.
We develop six tracker variations and conduct a thorough comparison of them on sequences covering various scenarios and motion complexities.
We release our implementations and detailed results to benefit the research community on event cameras: 
\url{https://github.com/gogojjh/ESVO_extension}.


\end{abstract}
\section{Introduction}
\label{sec:introduction}

Event cameras are novel and bio-inspired sensors.
Different from conventional frame cameras that capture images at a fixed rate, they asynchronously capture the per-pixel 
\textit{intensity changes} and output a stream of \textit{events}.
Each event is encoded with information, including the triggered time, pixel localization, and the sign of the intensity change.
As summarized in \cite{gallego2020event}, event cameras have high temporal resolution ($\mu s$-level), high dynamic range ($140$dB v.s. $60$dB of standard cameras),
and low power consumption.
These characterisics enable event cameras to have great potential for several computer vision and robotic tasks \eg,
high-speed motion estimation \cite{gallego2017event,bryner2019event} as well as feature tracking \cite{zhu2017event,gentil2020idol}  
and high dynamic range perception \cite{pan2019bringing,rebecq2019high}, 
which are difficult to frame cameras.
However, research on standard vision problems with event cameras is still preliminary. 
This is because event cameras work in a fundamentally different way from frame cameras, 
which measure \textit{intensity changes} asynchronously rather than the \textit{absolute intensity} (\ie, grayscale data) at a constant rate.
Thus, novel algorithms to process events must be investigated.

In this paper, we restrict the scope of literature to the SLAM problem using only event cameras, 
which does not include methods that use frame cameras \cite{vidal2018ultimate} or structured light \cite{martel2018active} 
to provide additional color/depth information.
Such a SLAM problem has been addressed step by step in scenarios with an increasing complexity.
Depending on the complexity of this problem, recent works can be categorized along with three axes:
\textit{1)} problem dimensionality: from individually handling the localization and 3D reconstruction subproblem \cite{gallego2017eventbased}
to solving the complete tracking-and-mapping problem \cite{rebecq2016evo};
\textit{2)} type of motion: estimating from constrained motions such as pure rotation or planar motion \cite{gallego2018unifying,liu2020globally,peng2020globally} 
to arbitrary 6-DoF motions \cite{zhou2021event}; 
\textit{3)} type of scenes: from artificial patterns \cite{mueggler2015continuous} to natural scenes \cite{rebecq2019events}.

Two event camera-based methods, \textit{EVO} \cite{rebecq2016evo} and \textit{ESVO} \cite{zhou2021event}, 
stand out as solving the SLAM problem in the most general setting (6-DoF motion and natural 3D scenes).
Different from the earlier work \cite{kim2016real} that needs to recover absolute intensity on a dedicated hardware (GPU), 
both EVO and ESVO directly directly run on a CPU in real-time.

EVO was proposed to solve the monocular event-based state estimation, 
which consists of a plane sweep-based mapper and a tracker that warps a semi-dense 3D map onto a binary event map (EM).
The tracker and mapper work in a parallel manner.
ESVO also follows this tracking-and-mapping philosophy to design the stereo event-based Visual Odometry (VO) system.
It exploits the novel time surface (TS), which encodes spatio-temporal constraints, to estimate a semi-dense depth map for each stereo observation pair. 
It utilizes the ``negative'' TS, which is aligned by the 3D map, to optimize pose parameters in tracking.
In summary, trackers in both EVO and ESVO resemble the similar frame-based paradigm \cite{forster2016svo}:
they use non-linear optimization to solve the 3D-2D alignment problem on image-type representations.
Although the EM and TS have been separately applied in above event-based VO systems,
the current literature has not offered comparative results of them.

The motivation of this paper is to address this deficiency by performing a comprehensive evaluation of the two representations. 
Furthermore, we enhance the original TS-based tracker by integrating it with the EM counterpart.
This tracker has a degeneracy evaluation module to determine which representations are used.
All implementations will be publicly released, which are developed from the open-source ESVO\footnote{\url{https://github.com/HKUST-Aerial-Robotics/ESVO}}.
We conduct experiments on simulated and real-world sequences, covering various scenarios from artificial patterns to natural scenes 
under motion in different complexities.
Besides the tracking accuracy, we also analyze the limitations of the EM and TS by presenting several failure cases.
We hope that our results and conclusions may reference researchers on event-based SLAM 
and indicate possible directions to improve the current state-of-the-art (SOTA) methods.

\section{Overview of Camera Tracking}
\label{sec:methodology}

\subsection{Formulation}
\label{sec:formulation}
The tracking module estimates 6-DoF poses of the event camera by taking the image-type event representations (also called event frames) as input.
It assumes that a semi-dense 3D map of the environment is given.
We denote the reference camera frame as $()^{r}$ and the current camera frame as $()^{c}$.
We use $M$ to indicate the template image in $()^{r}$ which is projected by the 3D map $\mathcal{M}$
and $\bar{I}$ to denote the current image.
As explained in the following sections, $\bar{I}$ is defined as the ``negative'' EM or TS.
The goal of the tracking is to find the transformation $T(\bm{\theta})$ that optimally aligns 
$\mathcal{M}$ onto $\bar{I}$. 
This ``alignemnt'' is done through the warping function:
\begin{equation}
    \begin{split}
        W(\mathbf{x},d;\bm{\theta})
        \doteq
        \pi\big(T(\bm{\theta}) \cdot \pi^{-1}(\mathbf{x}, d)\big),
        \ \ \ \mathbf{x}\in M,
    \end{split}
\end{equation}
where $\bm{\theta}$ are the pose parameters; 
$T(\bm{\theta})$ turns $\bm{\theta}$ into a transformation matrix from $()^{r}$ to $()^{c}$;
$\mathbf{x}=[u,v]$ is the localization of a valid pixel on the template image;
$\pi(\cdot)$ projects a 3D point onto the image plane, while 
$\pi^{-1}(\cdot)$ back-projects a pixel into 3D space with the known depth $d$.
We minimize the objective function to find the optimal $\bm\theta^{*}$ as
\begin{equation}
    \label{equ:objective_alignment}
    \underset{\bm{\theta}}{\arg\min}
    \sum_{\mathbf{x}\in M}
    \rho
    \Big(
        \bar{I}\big(W(\mathbf{x},d;\bm{\theta})\big)^{2}
    \Big),
\end{equation}
where $\bar{I}(\cdot)$ is the pixel value and $\rho(\cdot)$ is the robust loss.
Following ESVO, 
we reformulate Problem \eqref{equ:objective_alignment} using the forward compositional Lucas-Kanade method,
which iteratively refines the incremental pose parameters $\Delta\bm{\theta}$ by minimizing the objective as
\begin{equation}
    \label{equ:objective_forward_comp_LK}
    \begin{split}
        \underset{{\Delta\bm{\theta}}}{\arg\min}
        \sum_{\mathbf{x}\in M}
        \rho
        \Big(
            \bar{I}\big(W(W(\mathbf{x},d;\Delta\bm{\theta});\bm{\theta})\big)&^{2}
        \Big),        
    \end{split}
\end{equation}
where the warping function is updated at each iteration
\begin{equation}
    \begin{split}
        W(\mathbf{x},d;\bm{\theta})
        &\leftarrow
        W(W(\mathbf{x},d;\Delta\bm{\theta});\bm{\theta}).
    \end{split}
\end{equation}

The compositional approach is more efficient than the original Lucas-Kanade algorithm.
This is because the Jacobian $\frac{\partial W}{\partial\bm{\theta}}$ of the warping function is evaluated at a fixed point: $(\mathbf{x},\mathbf{0})$ 
which can be pre-computed.
Another reformulation is called the inverse compositional Lucas-Kanade algorithm, which is used in EVO.
It switches the roles of the template and current image.
In other words, we need to evaluate the gradient $\nabla M$ of $M$ as the part of the Jacobian, not of $\bar{I}$ in the forward approach.
But this paper considers the forward approach since it exploits the implicit ``slope'' feature of the TS.
Moreover, the definition of $\bm{\theta}$ varies according to different parameterization ways, including the Lie algebra \cite{barfoot2017state} or
Euler angles/Quaternions \cite{sola2017quaternion}/Cayley parameters \cite{cayley1846algebraic} for rotation with translation vectors.

\begin{figure}[t]
    \centering
    \subfigure[An example of the time surface which is triggered at every $10ms$]
    {
        \centering
        \includegraphics[width=0.465\linewidth]{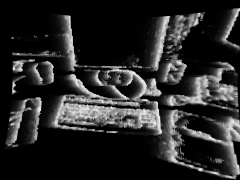}
        \label{fig:visual_TS}
    }        
    \hfill
    \subfigure[An example of the event map which aggregates $4000$ events]
    {
        \centering
        \includegraphics[width=0.465\linewidth]{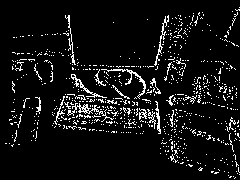}
        \label{fig:visual_EM}
    }    
    \caption{An example of the TS and EM.}
    \label{fig:visual_TS_EM}
\end{figure}

\subsection{Two Event Representations}
\label{sec:event_representation}

\subsubsection{Event Map}
Besides EVO, the binary EM has also been applied in visual-inertial odometry \cite{rebecq2017real} 
where FAST corners \cite{rosten2006machine} are directly extracted and tracked.
Constructing the EM is simple: a group of events within a temporal neighborhood is aggregated onto the image.
The pixel is set to $255$ where an event is fired; otherwise, it is set to zero.
Therefore, EMs are output in an asynchronous manner. 
But due to the data-driven nature of event cameras, the number of events in the group should be tuned for a specfic dataset.
Otherwise if the number is small, the frequency of producing EMs increases when the camera moves rapidly or scenes are high-contrast. 
This will raise the computational burden to the tracker.

\subsubsection{Time Surface}
The TS is also an image where each pixel stores a value.
Using an exponential decay kernel, TSs can emphasize recent events over past events. 
Given an arbitrary timestamp $t$, the value at a pixel $\mathbf{x}$
is defined as
\begin{equation}
    \label{equ:ts}
    \begin{split}
        I(\mathbf{x}, t)
        \doteq
        \exp\bigg(
        -\frac{t-t_{last}(\mathbf{x})}{\delta}
        \bigg),
    \end{split}
\end{equation}
where $t_{last}(\mathbf{x}) \leqslant t$ is the timestamp of the last event at $\mathbf{x}$ and 
$\delta$ is the constant decay rate parameter (\ie, $30ms$). 
This equation converts events into an image whose “intensity” is a function of the motion history at that location.
Larger values correspond to a more recent motion.
Online adjusting the value of $t$ can signal the creation of TSs synchronously or asynchronously.
In experiments, we output TSs in a synchronous way.
For convenient visualization and processing, each TS is rescaled from $[0, 1]$ to the range $[0, 255]$.

\subsubsection{Summary}
We summarize the pros and cons of these representations from recent literature \cite{gallego2020event}.
Generally, EMs implicitly represent the edge map since events are mostly triggered by edge patterns. 
They are the universal data structure compatible with conventional computer vision.
Also, generating an EM is very fast (\ie, $<0.5ms$).
Nevertheless, EMs are highly sensitive to motion blur if the number of aggregating events is not set well.
In contrast, TSs are more informative than EMs even though they require around $5-10ms$ for the synthesis.
In experiments, we will further identify several limitations of both EMs and TSs in challenging scenarios.

The tracker operates the ``negative'' event frames \ie, $\bar{I}(\mathbf{x})=255-I(\mathbf{x})$.
This is based on a crucial observation: the objective \eqref{equ:objective_forward_comp_LK} becomes minimum if the template image is ``perfectly'' aligned on dark areas in the current frame.
An example can be seen in Fig. \ref{fig:tracking_deg_reproj_em}.
Specifically, the negative TSs can be interpreted as an anisotropic distance field as presented in edge-based VO \cite{zhou2018canny}.
To enlarge the convergence basin's width in optimization, a Gaussian blur (kernel size = $5$ pixels) is applied to negative event frames.

\subsection{Enhanced Tracker with Degenearcy Check}
\label{sec:degeneracy_check}
The synchronous pose output of the TS-based tracker is beneficial to 
sensor fusion like the visual-inertial odometry or other robotic applications (\eg, drone racing).
However, unlike the EM that aggregates a fixed number of events, 
the TS becomes unreliable if few events are triggered (see Section \ref{sec:experiment}).
This issue frequently occurs where event cameras are working in textureless scenarios or relatively static to scenes. 
The ``jump'' in motion estimates is inevitable due to unconstrained degrees of freedom.

In this paper, we enhance the original TS-based tracker in robustness by taking EMs as the backup representation.
The proposed tracker is based on a test that online evaluates the degeneracy of the optimization problem \eqref{equ:objective_forward_comp_LK}.
Inspired by \cite{zhang2016degeneracy}, we use the \textit{factor} $\lambda$ as the degeneracy quantification metric.
$\lambda$ is defined as the minimum eigenvalue of the \textit{Hessian matrix} 
which is computed by linearizing \eqref{equ:objective_forward_comp_LK} as a normal equation: $\mathbf{H}\Delta\bm{\theta}=\mathbf{g}$.
We take both TSs and EMs as input, 
but first utilize the TS-based tracker.
Before optimization, we compute $\lambda$ and check if it is smaller than a threshold $\lambda_{th}$.
If $\lambda<\lambda_{th}$, problem \eqref{equ:objective_forward_comp_LK} may degenerate, 
and we use the EM-based tracker to optimize pose parameters. 
The EM is constructed by aggregating recent $4000$ events.

\begin{figure}[t]
    \subfigure[\textit{simu\_office}]
    {\centering\includegraphics[width=0.315\linewidth]{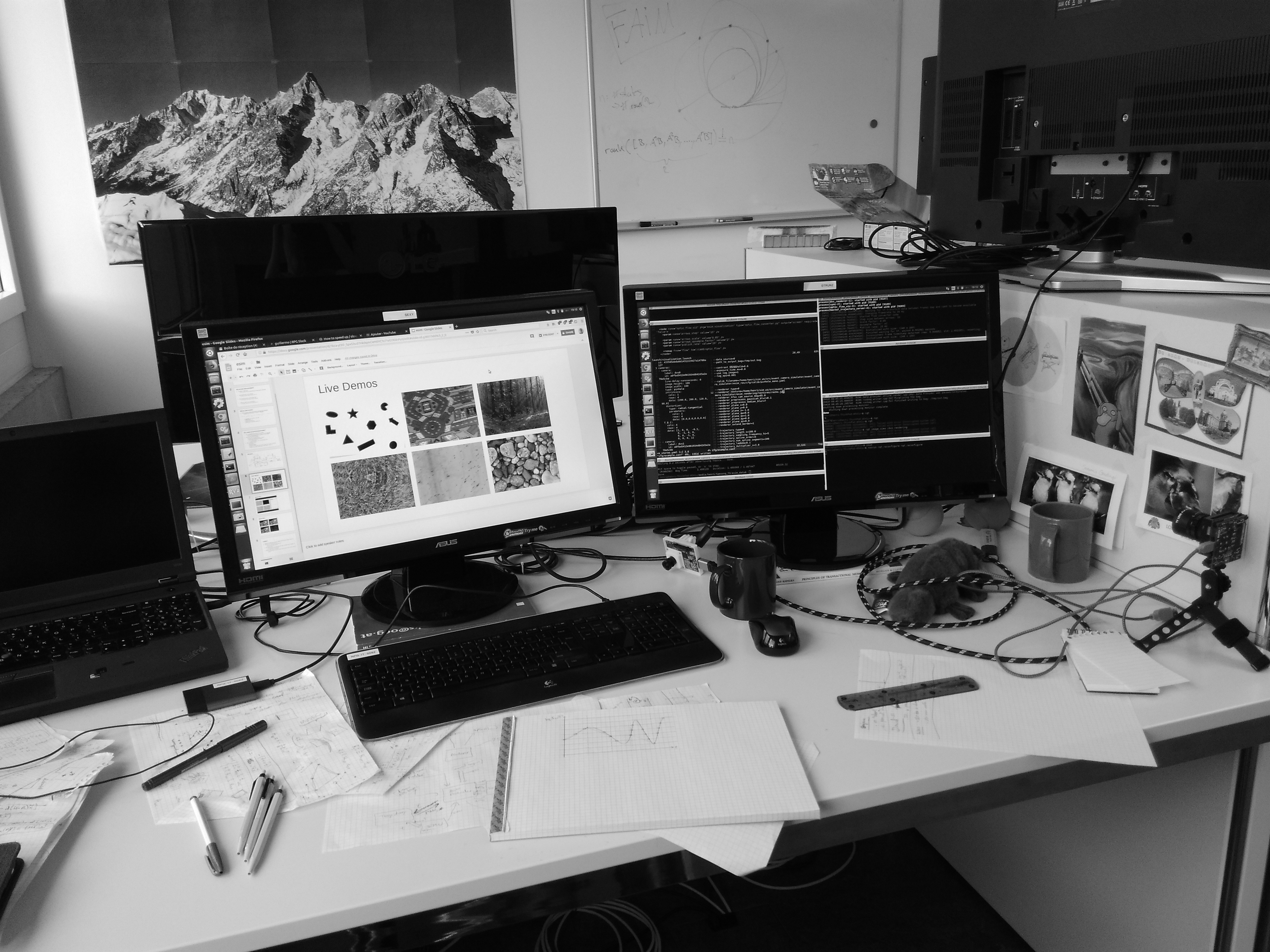}
    \label{fig:scene_simu_office}}
    \hfill\subfigure[\textit{simu\_poster}]
    {\centering\includegraphics[width=0.315\linewidth]{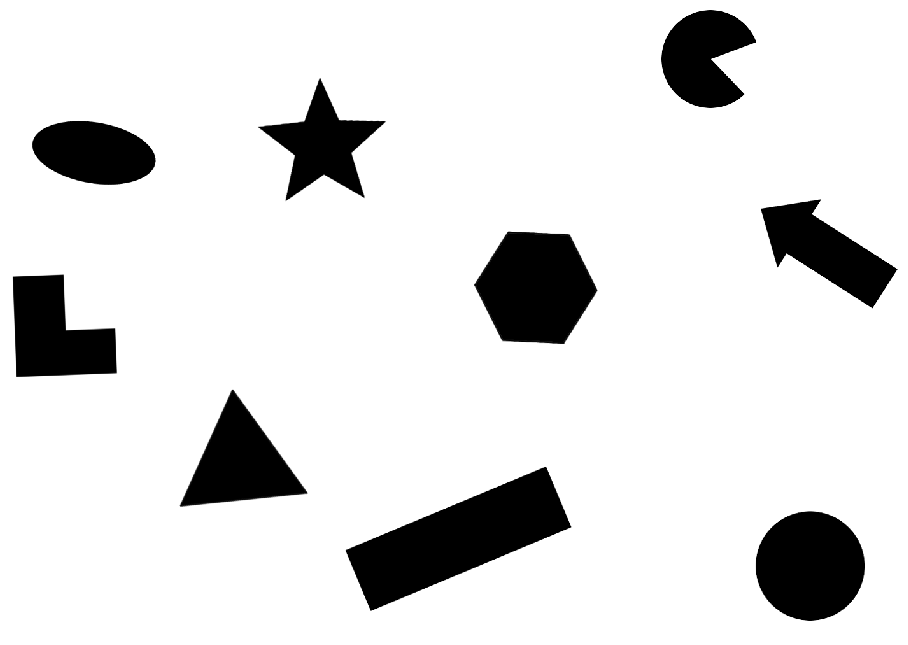}
    \label{fig:scene_simu_poster}}
    \hfill\subfigure[\textit{simu\_checkerboard}]
    {\centering\includegraphics[width=0.315\linewidth]{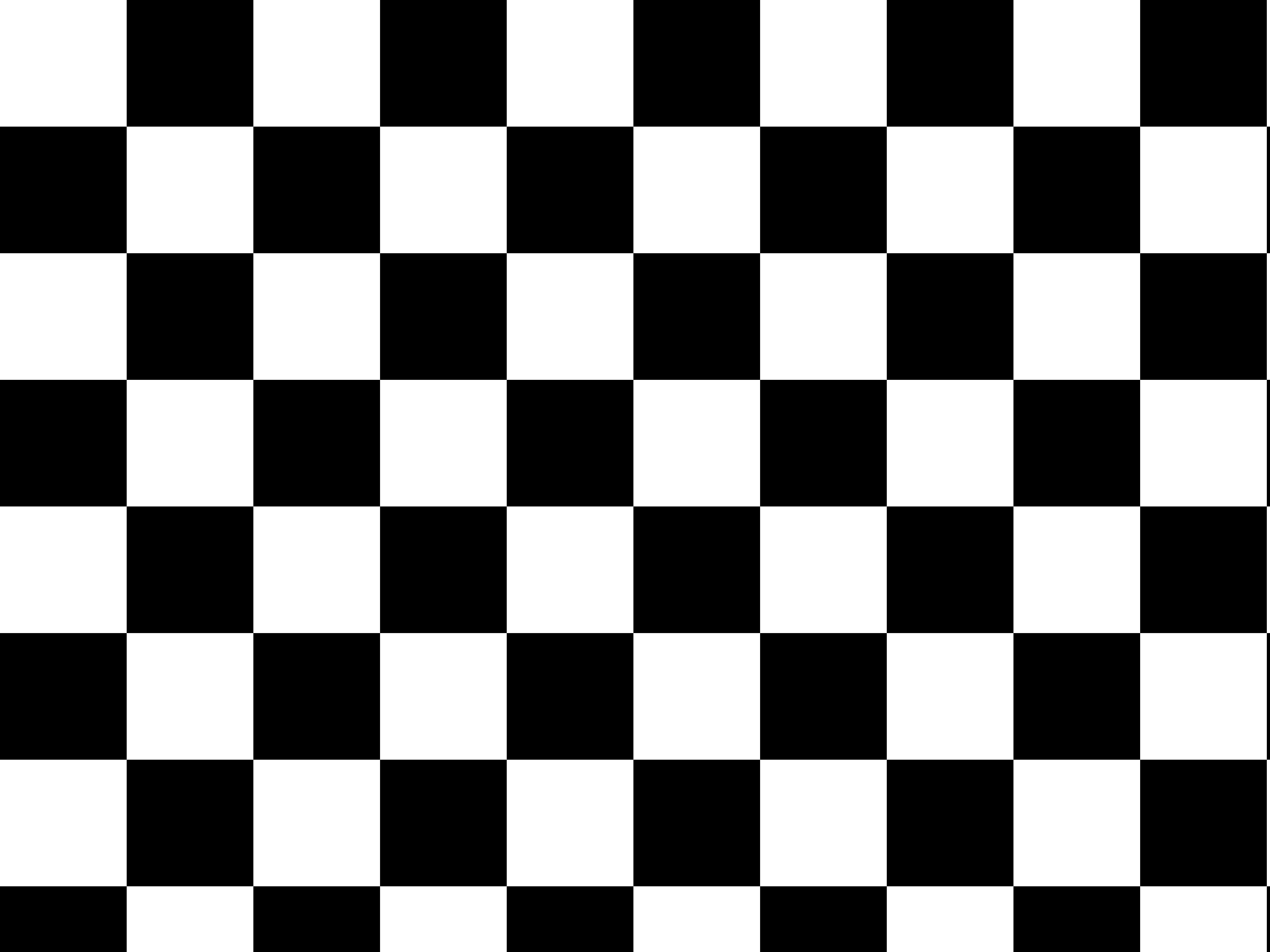}
    \label{fig:scene_simu_checkerboard}}
    \subfigure[\textit{rpg\_bin}]
    {\centering\includegraphics[width=0.315\linewidth]{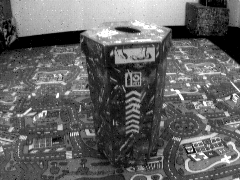}
    \label{fig:scene_rpg_bin}}    
    \hfill\subfigure[\textit{rpg\_box}]
    {\centering\includegraphics[width=0.315\linewidth]{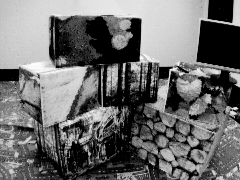}
    \label{fig:scene_rpg_box}}    
    \hfill\subfigure[\textit{rpg\_desk}]
    {\centering\includegraphics[width=0.315\linewidth]{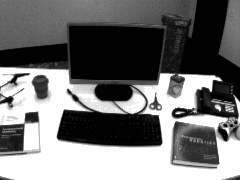}
    \label{fig:scene_rpg_desk}}    
    \subfigure[\textit{rpg\_monitor}]
    {\centering\includegraphics[width=0.315\linewidth]{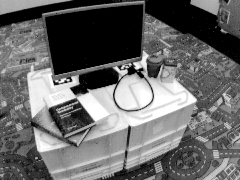}
    \label{fig:scene_rpg_monitor}}    
    \hfill\subfigure[\textit{upenn\_flying1}]
    {\centering\includegraphics[width=0.315\linewidth]{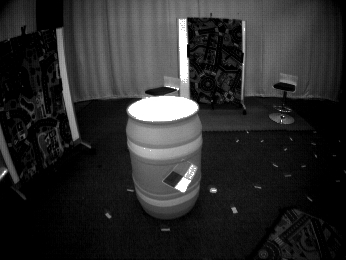}
    \label{fig:scene_upenn_indoor_flying1}}
    \hfill\subfigure[\textit{upenn\_flying3}]
    {\centering\includegraphics[width=0.315\linewidth]{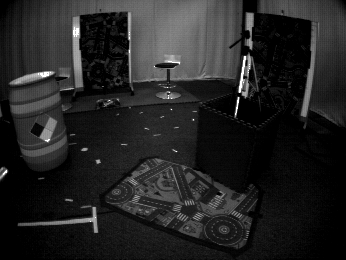}
    \label{fig:scene_upenn_indoor_flying3}}    
    \caption{Scene images of $\textit{simu}$, $\textit{rpg}$, and $\textit{upenn}$ sequences. 
    The camera resolutions are $346\times 260$, $240\times 180$, and $346\times 260$, respectively.}
    \label{fig:scene_image}
\end{figure}

\section{Experiment}
\label{sec:experiment}

\begin{table*}[t]
    \begin{center}
        \renewcommand\arraystretch{1.0}
        \renewcommand\tabcolsep{10pt}
        \begin{tabular}{lcccccc}
            \toprule[0.03cm]
            Sequence (setting different $\lambda_{th}$) & $10$ & $31$ & $100$ & $158$ & $251$ & $400$ \\ 
            \toprule[0.03cm]
            \textit{simu\_office\_planar}             & $5.5$ & $\textbf{4.9}$ & $\textbf{4.7}$ & $5.1$ & $6.8$ & $8.2$ \\
            \textit{simu\_poster\_planar}             & $4.6$ & $4.6$ & $4.6$ & $4.6$ & $\textbf{4.5}$ & $\textbf{4.4}$ \\
            \textit{simu\_checkerboard\_planar}       & $\textbf{4.7}$ & $\textbf{4.7}$ & $5.6$ & $5.3$ & $7.5$ & $9.2$ \\
            \midrule
            \textit{simu\_office\_6DoF}               & $\textbf{17.4}$ & $\textbf{18.7}$ & $22.0$ & $33.7$ & $42.8$ & $36.6$ \\
            \textit{simu\_poster\_6DoF}               & $\textbf{16.9}$ & $17.3$ & $\textbf{17.1}$ & $19.7$ & $19.0$ & $19.4$ \\
            \textit{simu\_checkerboard\_6DoF}         & $\textbf{26.1}$ & $\textbf{28.1}$ & $30.3$ & $32.2$ & $32.6$ & $32.1$ \\
            \midrule
            \textit{rpg\_bin\_6DoF}                   & $3.6$ & $3.8$ & $3.6$ & $\textbf{3.4}$ & $\textbf{3.4}$ & $5.2$ \\
            \textit{rpg\_box\_6DoF}                   & $\textbf{6.7}$ & $\textbf{7.1}$ & $42.0$ & $116.8$ & $126.7$ & $125.3$ \\
            \textit{rpg\_desk\_6DoF}                  & $3.8$ & $3.8$ & $3.6$ & $\textbf{3.5}$ & $\textbf{3.5}$ & $16.2$ \\
            \textit{rpg\_monitor\_6DoF}               & $7.3$ & $\textbf{7.0}$ & $7.4$ & $7.2$ & $7.2$ & $\textbf{6.6}$ \\
            \midrule
            \textit{upenn\_indoor\_flying1\_6DoF}     & $16.1$ & $14.8$ & $15.0$ & $\textbf{14.5}$ & $17.1$ & $\textbf{14.0}$ \\
            \textit{upenn\_indoor\_flying3\_6DoF}     & $10.9$ & $10.9$  & $11.0$  & $10.1$  & $\textbf{8.9}$  & $\textbf{7.3}$  \\
            \midrule
            Count of the best score                   & $5$ & $\textbf{6}$ & $2$ & $3$ & $4$ & $4$ \\
            \bottomrule
        \end{tabular}
    \end{center}
    \caption{Setting different $\lambda_{th}$, the TSEM-based trackers are separately tested under $10$-trials. 
             Mean ATE in translation $[cm]$ is shown. The first two lowest errors are marked as bold.}
    \label{tab:ATE_diff_lambda}
\end{table*}

\begin{figure*}[t]
    \centering
    \subfigure[]
    {\centering\includegraphics[width=0.155\linewidth]{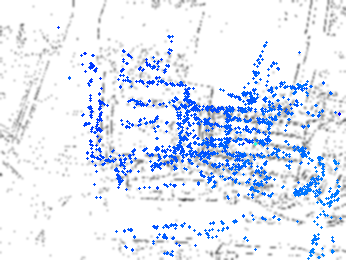}
     \label{fig:EM_tracking_inconsistency_case2_1}}
    \hfill
    \subfigure[]
    {\centering\includegraphics[width=0.155\linewidth]{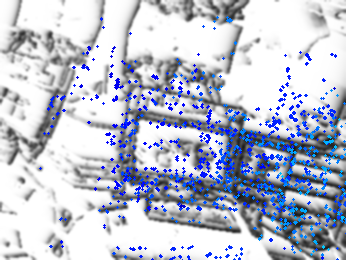}
     \label{fig:EM_tracking_inconsistency_case2_2}}
    \hfill
    \subfigure[]
    {\centering\includegraphics[width=0.155\linewidth]{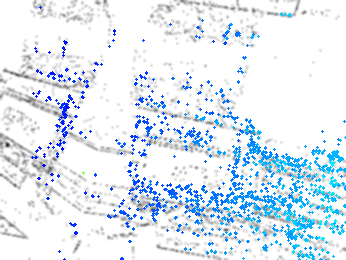}
     \label{fig:EM_tracking_inconsistency_case2_3}}
    \hfill
    \subfigure[]
    {\centering\includegraphics[width=0.155\linewidth]{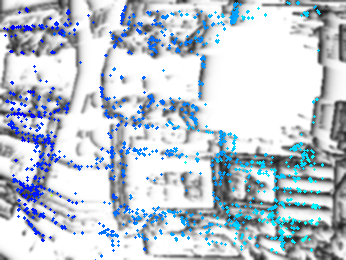}
     \label{fig:EM_tracking_inconsistency_case2_4}}
    \hfill
    \subfigure[]
    {\centering\includegraphics[width=0.155\linewidth]{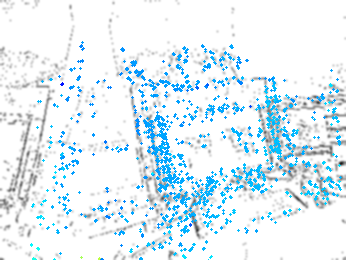}
     \label{fig:EM_tracking_inconsistency_case2_5}}
    \hfill
    \subfigure[]
    {\centering\includegraphics[width=0.155\linewidth]{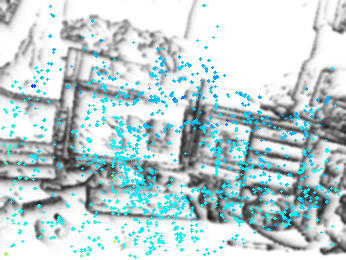}
     \label{fig:EM_tracking_inconsistency_case2_6}}

    \hfill
    \subfigure[]
    {\centering\includegraphics[width=0.155\linewidth]{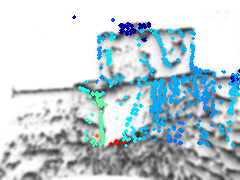}
     \label{fig:EM_tracking_inconsistency_case1_1}}
    \hfill
    \subfigure[]
    {\centering\includegraphics[width=0.155\linewidth]{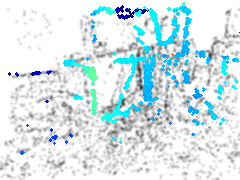}
     \label{fig:EM_tracking_inconsistency_case1_2}}
    \hfill
    \subfigure[]
    {\centering\includegraphics[width=0.155\linewidth]{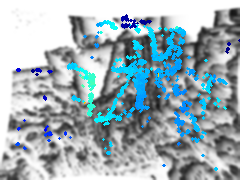}
     \label{fig:EM_tracking_inconsistency_case1_3}}
    \hfill
    \subfigure[]
    {\centering\includegraphics[width=0.155\linewidth]{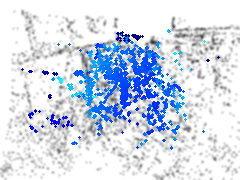}
     \label{fig:EM_tracking_inconsistency_case1_4}}
    \hfill
    \subfigure[]
    {\centering\includegraphics[width=0.155\linewidth]{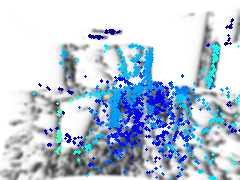}
     \label{fig:EM_tracking_inconsistency_case1_5}}
    \hfill
    \subfigure[]
    {\centering\includegraphics[width=0.155\linewidth]{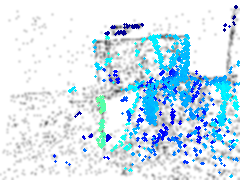}
     \label{fig:EM_tracking_inconsistency_case1_6}}
    \caption{Consecutive tracking results on $\textit{simu\_office\_6DoF}$ (\ref{fig:EM_tracking_inconsistency_case2_1} - \ref{fig:EM_tracking_inconsistency_case2_6}) 
            and $\textit{rpg\_box}$ (\ref{fig:EM_tracking_inconsistency_case1_1} - \ref{fig:EM_tracking_inconsistency_case1_6}) respectively. Due to the large $\lambda_{th}=158$, 
            TSs and EMs are frequently changed for tracking.}
    \label{fig:TSEM_tracking_inconsistency_case}
\end{figure*}
    \subsection{Implementation Details}
We use the Eigen library \cite{guennebaud2010eigen} to solve the nonlinear optimization problem.
The tracker requires a known 3D map, which is computed by the mapping module of ESVO in real-time.
We use two public datasets, and the simulator \cite{rebecq2018esim} to generate sequences for evaluation.
All data were collected with stereo event cameras, but the motion and scenes present various difficulties.
The simulator provides synthetic data with a planar structure and an ``ideal'', noise-free camera model.
We collect sequences (\textit{simu\_X}) using stereo event cameras before a wall with various backgrounds.
We category the collected simulated sequences into two classes according to the motion complexity: 
\textit{1)} slow ($\approx 0.3m/s$) and planar motion; \textit{2)} fast ($\approx 1.0m/s$) and 6-DoF motion.
Real-world data in \cite{zhou2021event} (\textit{rpg\_X}) were collected in an office, 
while that in \cite{zhu2018multivehicle} (\textit{upenn\_X}) were collected by mounting cameras on a flying drone in a capacious indoor area.
Note that the ``\textit{X}'' indicates the type of scenes and motions.
The ground-truth camera poses are provided.
Fig. \ref{fig:scene_image} shows the scene images.
In experiments, the algorithm is executed on a desktop with an i7 CPU@4.20 GHz. 

    \subsection{Setting the Proper $\lambda_{th}$}
\label{sec:exp_diff_lambda}

\begin{figure}[t]
    \centering
    \subfigure[The value of the degeneracy factor $\lambda$ on different sequences]
    {\centering\includegraphics[width=0.99\linewidth]{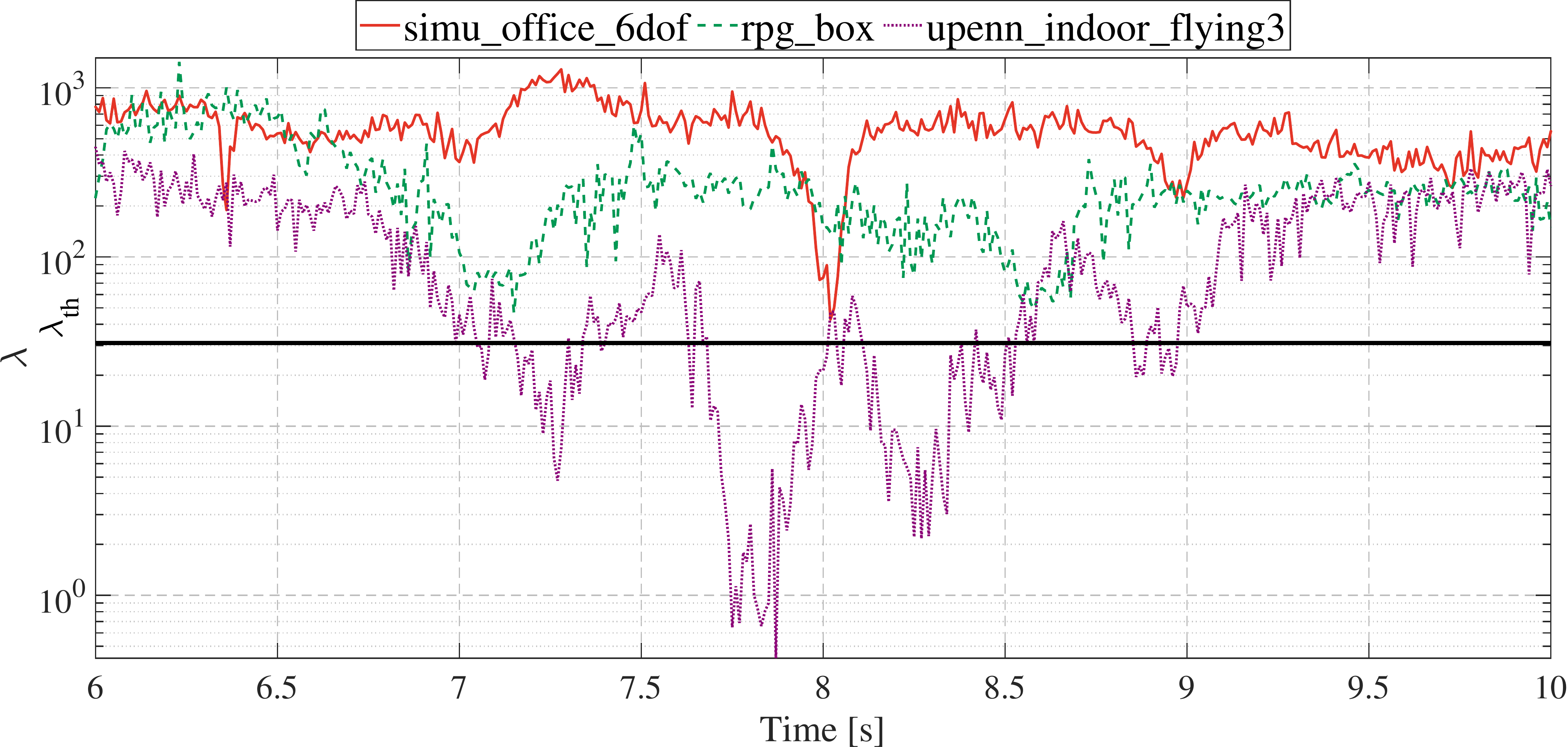}
     \label{fig:tracking_deg_factor}}
    \subfigure[The 3D map is warped on the negative TS.]
    {\centering\includegraphics[width=0.47\linewidth]{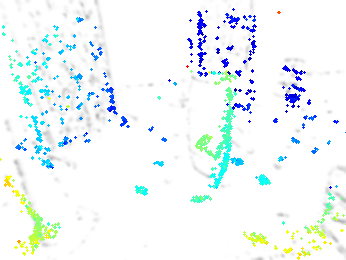}
     \label{fig:tracking_deg_reproj_ts}}        
    \hfill
    \subfigure[The 3D map is warped on the negative $\text{EM}$ with $4000$ events.]
    {\centering\includegraphics[width=0.47\linewidth]{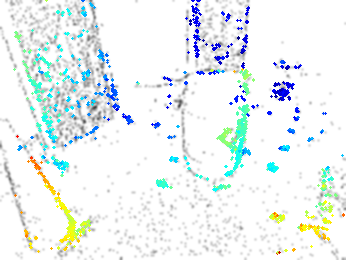}
     \label{fig:tracking_deg_reproj_em}}    
    \caption{We plot the values of $\lambda$ on three sequences in (a).
    At the time of $8.2s$ of \textit{upenn\_indoor\_flying1} (b) where the TS-based tracker degenerates, 
    (c) the 3D map is well aligned on ``dark areas'' of the negative EM with sufficient constraints.}
    \label{fig:tracking_deg}
\end{figure}

The degeneracy threshold $\lambda_{th}$ is a hard-tuned parameter.
It implcitly indicates the frequency of using the backup EM representation for tracking.
The larger $\lambda_{th}$, the more frequent usage of EMs. 
To study this effect on the tracking accuracy, we first conduct separate tests on the TSEM-based tracker by setting different $\lambda_{th}$.
Note that the TSEM-based tracker is the enhanced tracker with the degeneracy check in Section \ref{sec:degeneracy_check}.
The EM aggregate recent $4000$ events. 
This tracker is denoted by ``TS$\text{EM}_{4000}$'' in Section \ref{sec:tracking_evaluation}.
According to Fig. \ref{fig:tracking_deg_factor}, we increase the value of $\lambda_{th}$ at an exponential scale: 
$10^{\alpha},\ \alpha\in\{1,1.5,2,2.2,2.4,2.6\}$.
Each test is performed under 10-trials. The pose accuracy using the mean Abolute Trajectory Error (ATE) in translation \cite{zhang2018tutorial} 
w.r.t. the ground turth is reported in Table \ref{tab:ATE_diff_lambda}.
The error increase significantly on $\textit{simu\_office\_6DoF}$ and $\textit{rpg\_box}$ when $\lambda_{th}$ is large (\eg, $100$).
On these sequences, we observe that the tracker frequently switch these two representations for tracking 
and change the objective function \eqref{equ:objective_forward_comp_LK} in optimization for each input frame.
We explain that this may easily cause the inconsistency in optimization, and thus fail the tracker.
Fig. \ref{fig:TSEM_tracking_inconsistency_case} show consecutive tracking results 
at the begining of $\textit{simu\_office\_6DoF}$ and $\textit{rpg\_box}$ respectively.
Here, the tacker sets $\lambda_{th}=158$.

From Table \ref{tab:ATE_diff_lambda}, each tracker is counted with the number of getting the best score.
We find that setting $\lambda_{th}=31$ obtains stable and promissing results. 
Thus, the following experiments use this threshold.


\begin{figure}[t]
    \centering
    \subfigure[Failure case 1]
    {\centering\includegraphics[width=0.31\linewidth]{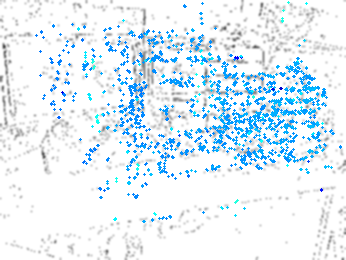}
     \label{fig:EM_tracking_fail_simu_office_6dof}}
    \hfill
    \subfigure[Failure case 2]
    {\centering\includegraphics[width=0.31\linewidth]{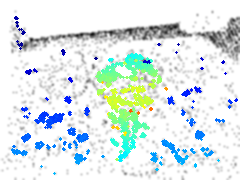}
     \label{fig:EM_tracking_fail_rpg_bin}}
    \hfill
    \subfigure[Failure case 3]
    {\centering\includegraphics[width=0.31\linewidth]{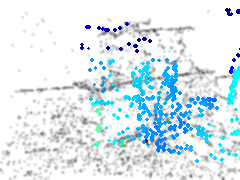}
     \label{fig:EM_tracking_fail_rpg_box}}    
    \caption{Failure examples of the EM-based tracker that aggregates recent $3000$ events. 
    Sequences from left to right are: \textit{simu\_office\_6DoF}, \textit{rpg\_bin}, and \textit{rpg\_box}.}
    \label{fig:EM_tracking_fail}
\end{figure}
 
\begin{table*}[t]
    \begin{center}
        \renewcommand\arraystretch{1.0}
        \renewcommand\tabcolsep{10pt}
        \begin{tabular}{lcccccc}
            \toprule
            Sequence & TS & $\text{EM}_{2000}$ & $\text{EM}_{3000}$ & $\text{EM}_{4000}$ & $\text{EM}_{5000}$ & TS$\text{EM}_{4000}\ (\lambda_{th}=31)$  \\ 
            \toprule
            \textit{simu\_office\_planar}                   & $4.7$ & $4.0$ & $\textbf{3.9}$ & $\textbf{3.7}$ & $4.1$ & $4.9$ \\
            \textit{simu\_poster\_planar}                   & $4.7$ & $\textbf{3.7}$ & $\textbf{4.3}$ & $4.6$ & $5.0$ & $4.6$ \\
            \textit{simu\_checkerboard\_planar}             & $4.2$ & $2.9$ & $\textbf{2.2}$ & $\textbf{2.3}$ & $2.4$ & $4.7$ \\
            \midrule
            \textit{simu\_office\_6DoF}               & $\textbf{9.1}$ & $25.3$ & $21.0$ & $16.6$ & $15.8$ & $18.7$ \\
            \textit{simu\_poster\_6DoF}               & $18.2$ & $\textbf{15.4}$ & $\textbf{16.3}$ & $16.8$ & $17.4$ & $17.3$ \\
            \textit{simu\_checkerboard\_6DoF}         & $23.0$ & $17.0$ & $\textbf{14.0}$ & $15.1$ & $\textbf{13.4}$ & $28.1$ \\
            \midrule
            \textit{rpg\_bin\_6DoF}                         & $\textbf{3.4}$ & $22.4$ & $16.6$ & $8.0$ & $14.1$ & $3.8$ \\
            \textit{rpg\_box\_6DoF}                         & $\textbf{6.5}$ & $\textbf{5.3}$ & $17.1$ & $13.7$ & $9.8$ & $7.1$ \\
            \textit{rpg\_desk\_6DoF}                        & $3.4$ & $\textbf{2.9}$ & $3.3$ & $3.2$ & $\textbf{2.9}$ & $3.8$ \\
            \textit{rpg\_monitor\_6DoF}                     & $7.2$ & $\textbf{5.3}$ & $\textbf{5.2}$ & $7.4$ & $7.3$ & $7.0$ \\
            \midrule
            \textit{upenn\_indoor\_flying1\_6DoF}           & $18.5$ & $22.0$ & $16.7$ & $\textbf{16.0}$ & $22.1$ & $\textbf{14.8}$ \\
            \textit{upenn\_indoor\_flying3\_6DoF}           & $20.9$ & $\textbf{10.8}$ & $11.9$ & $14.0$ & $15.0$ & $\textbf{10.9}$ \\
            \bottomrule
        \end{tabular}
    \end{center}
    \caption{Mean ATE in translation $[cm]$ under $10$-trials. The first two lowest errors are marked as bold.}
    \label{tab:ATE}
\end{table*}

\begin{figure*}[t]
    \centering\includegraphics[width=0.995\linewidth]{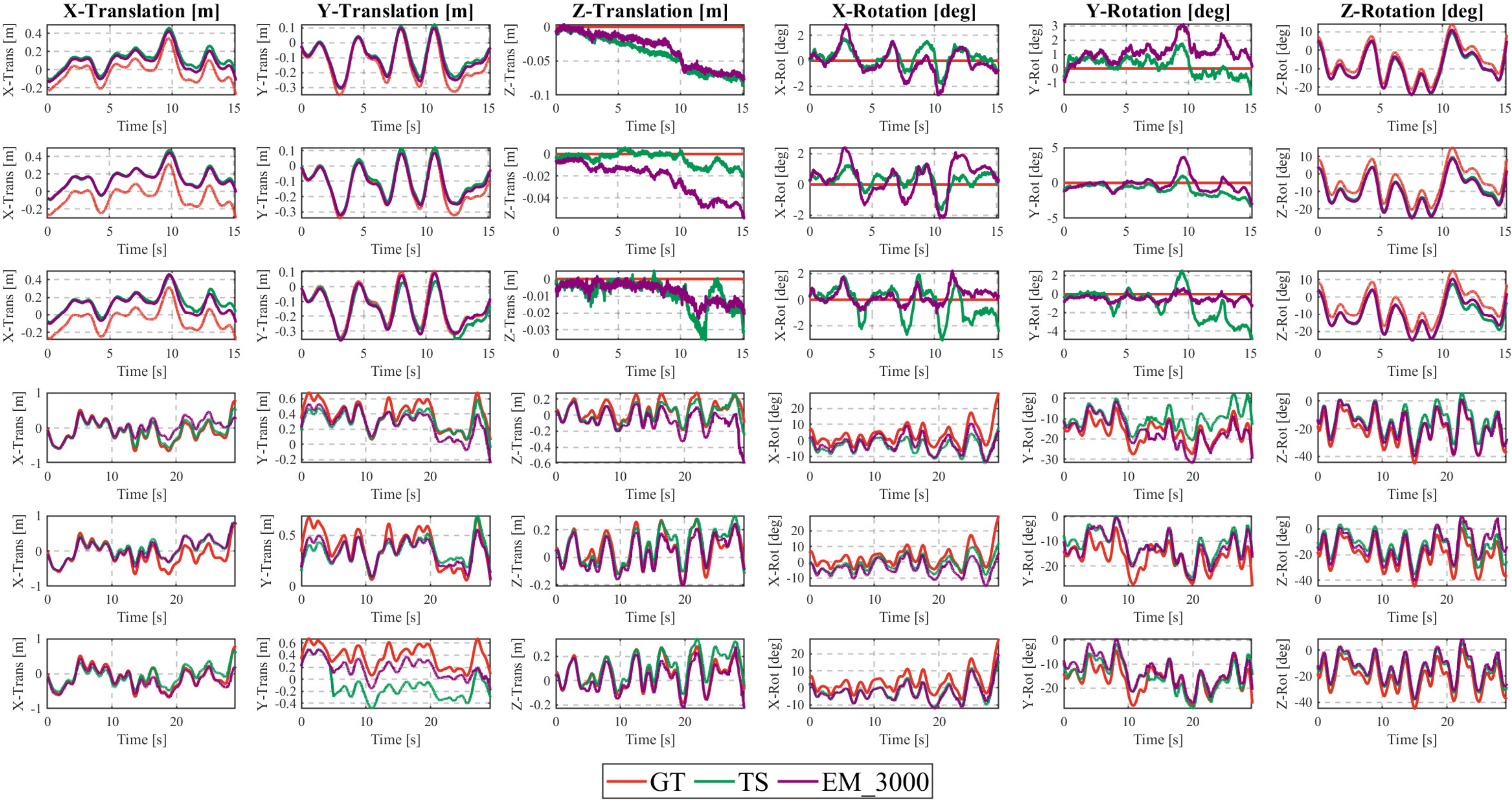}
    \caption{6-DoF poses comparison of the TS-based (``TS'') and EM-based tracker with the highest accuracy (``$\text{EM}_{3000}$'') against 
    the ground truth (``GT'') on \textbf{simulated\ sequences}.
    Each row corresponds to a different sequence: 
    \textit{simu\_office\_planar}, \textit{simu\_poster\_planar}, \textit{simu\_checkerboard\_planar}, \textit{simu\_office\_6DoF}, \textit{simu\_poster\_6DoF}, \textit{simu\_checkerboard\_6DoF}.
    These sequences are captured with stereo event cameras under random motions.}
    \label{fig:simu_dof_plot}
\end{figure*}

\begin{figure*}[t]
    \centering\includegraphics[width=0.995\linewidth]{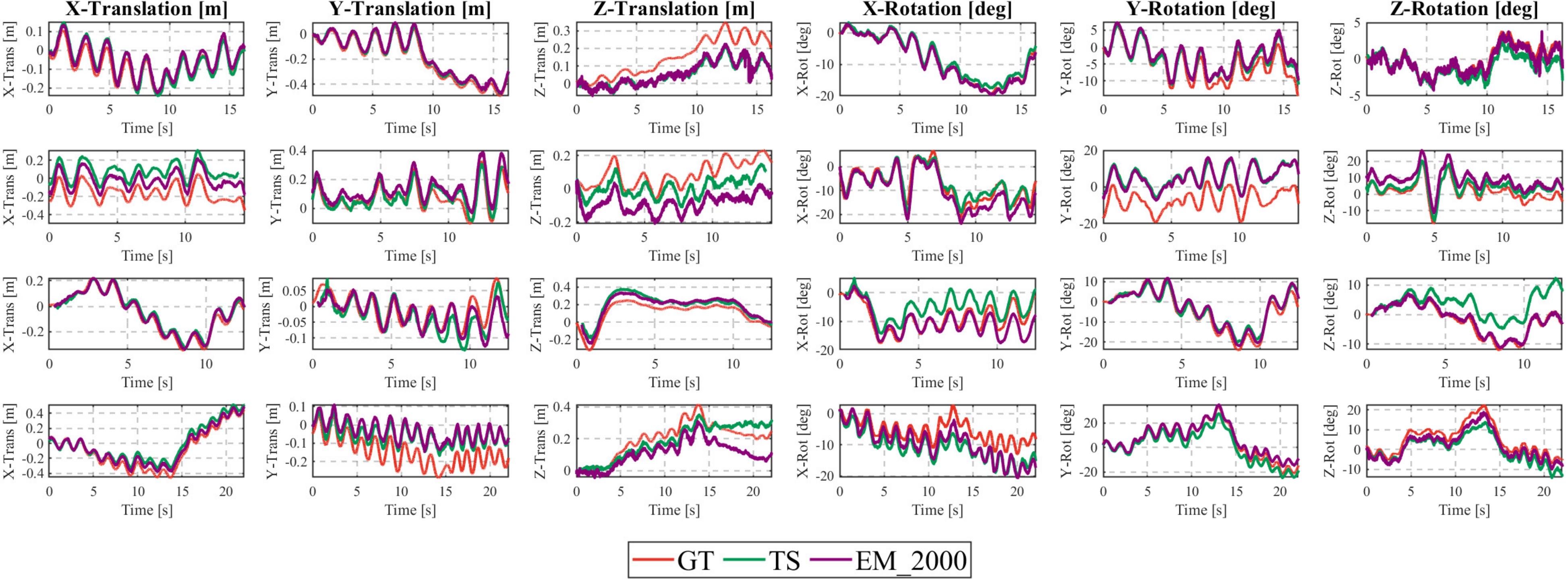}
    \caption{6-DoF poses comparison of the TS-based (``TS'') and EM-based tracker with the highest accuracy (``$\text{EM}_{2000}$'') against 
    the ground truth (``GT'') on \textbf{rpg\ sequences}.
    Each row corresponds to a different sequence: 
    \textit{rpg\_bin}, \textit{rpg\_box}, \textit{rpg\_desk}, \textit{rpg\_monitor}.
    These sequences are captured with a handheld stereo rig moving under a locally loopy behavior.}
    \label{fig:rpg_dof_plot}
\end{figure*}

\begin{figure*}[t]
    \centering\includegraphics[width=0.995\linewidth]{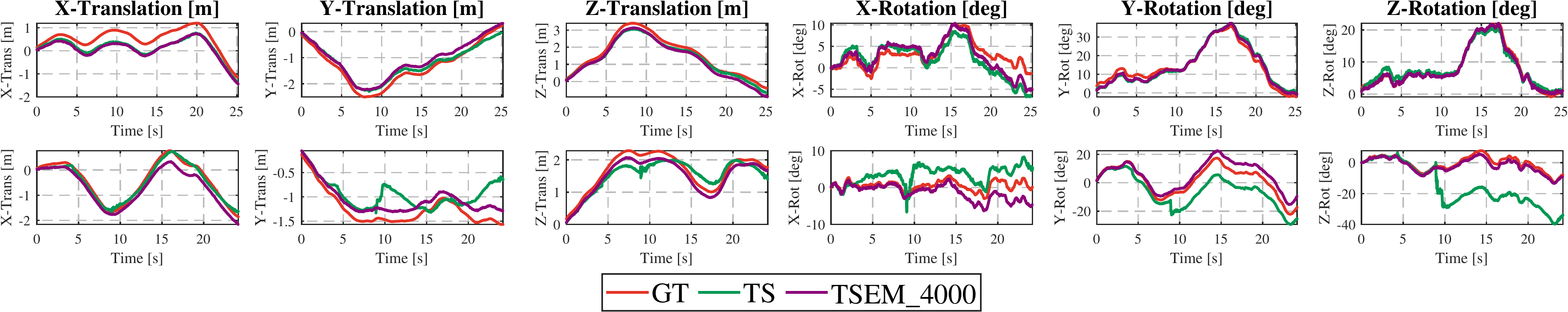}
    \caption{6-DoF poses comparison of the TS-based (``TS'') and TSEM-based tracker (``$\text{TSEM}_{4000}$'') against the ground truth (``GT'') on \textbf{upenn\ sequences}.
    Each row corresponds to a different sequence: \textit{upenn\_indoor\_flying1}, \textit{upenn\_indoor\_flying3}.
    These sequences are captured with a stereo rig mounted on a drone.}
    \label{fig:upenn_dof_plot}
\end{figure*}


    \subsection{Tracking Evaluation}
\label{sec:tracking_evaluation}

We implement six tracker variations according to the used representations.
They are denoted by ``TS'', ``$\text{EM}_{2000}$'', ``$\text{EM}_{3000}$'', ``$\text{EM}_{4000}$'', ``$\text{EM}_{5000}$'', and ``TS$\text{EM}_{4000}$'' respectively.
The subscript of ``EM'' indicates the number of events to generate an EM. 
``TS$\text{EM}_{4000}$'' is the enhanced tracker with the degeneracy check.
To evaluate their performance, we conduct $10$-trial SLAM tests on each sequence.
The mean ATE is reported in Table \ref{tab:ATE}.

All trackers achieve similar accuracy on the simplest sequences (\textit{simu\_X\_planar}), 
while they show worse accuracy on more complex sequences.
We cannot directly determine which representations are better for tracking according to the accuracy in Table \ref{tab:ATE} 
since all trackers may have large errors on specific sequences.
We should analyze the results by considering the characteristics of scenes and motion.
The TS-based tracker outperform others on \textit{simu\_office\_6DoF}, \textit{rpg\_bin}, and \textit{rpg\_box},
where high-contrast textures such as lines and rectangles are presented (see Fig. \ref{fig:scene_image}). 
In contrast, we observe that EM-based trackers reach another local minima which is far from the groundtruth during the optimization process
and thus result in low accuracy.
Fig. \ref{fig:EM_tracking_fail} illustrates three failure cases.

\begin{table}[t]
    \begin{center}
        \renewcommand\arraystretch{1.0}
        \renewcommand\tabcolsep{3pt}
        \begin{tabular}{cccccc}
            \toprule
            TS & $\text{EM}_{2000}$ & $\text{EM}_{3000}$ & $\text{EM}_{4000}$ & $\text{EM}_{5000}$ & TS$\text{EM}_{4000}$\\
            \toprule
            $8\pm5$ & $8\pm6$ & $7\pm4$ & $7\pm4$ & $7\pm4$& $7\pm3$ \\
            \bottomrule
        \end{tabular}
    \end{center}
    \caption{Average computation time [$ms$] of trackers.}
    \label{tab:latency}
\end{table}

\begin{figure}[t]
    \centering\includegraphics[width=0.995\linewidth]{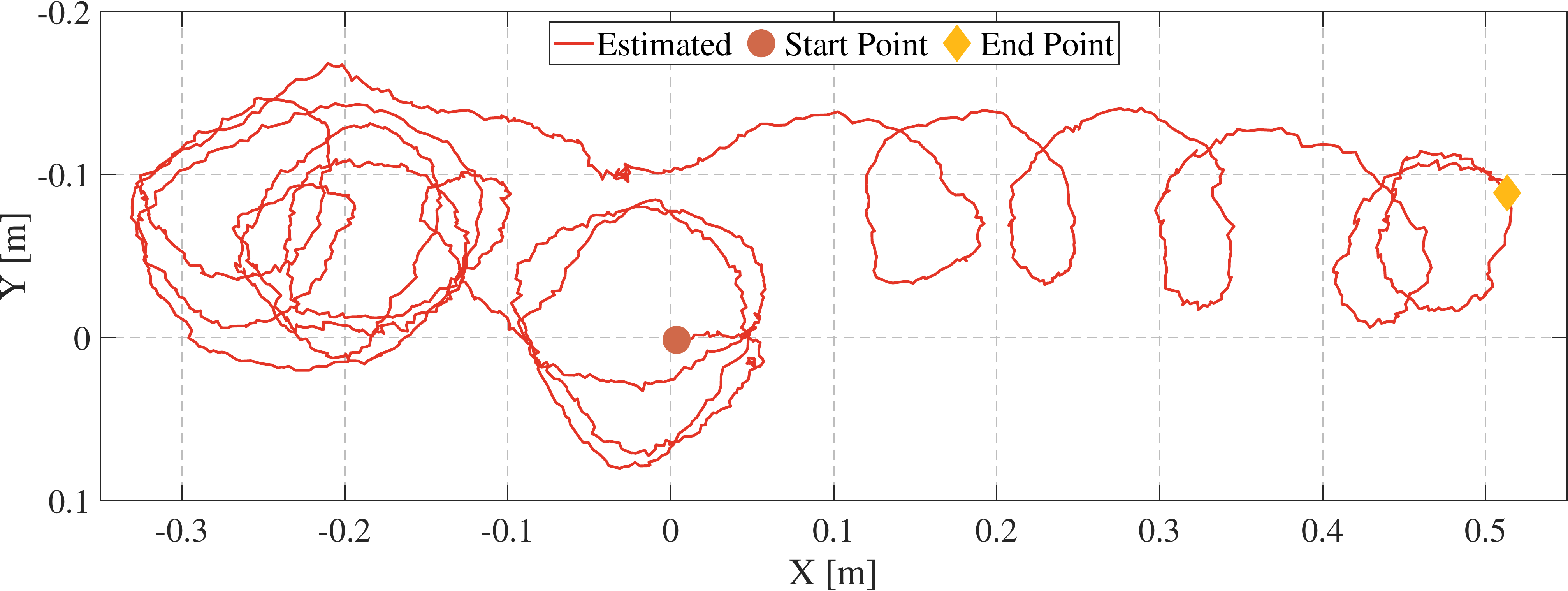}
    \caption{TS's trajectory on $\textit{rpg\_monitor}$.}
    \label{fig:TS_traj_rpg_monitor}
\end{figure}

On the other hand, EM-based trackers consistently have better performance than their TS-based counterpart on \textit{upenn\_indoor\_flying3}.
We note that few events are generated when the drone stops and hovers at some time.
This causes the generation of unreliable TSs which are triggered at a constant rate.
For the tracker, not enough constraints (\ie, ``dark areas'' on TSs) make the optimization degenerate or fail.
An example is shown in Fig. \ref{fig:tracking_deg}, 
with the plot displaying the value of the \textit{degeneracy factor} $\lambda$ on sequences: \textit{simu\_office\_6DoF}, \textit{rpg\_box}, and \textit{upenn\_indoor\_flying3}.
Regarding \textit{upenn\_indoor\_flying3}, at the time of $8.2s$, the TS-based tracker degenerates due to the unreliable TS (see Fig. \ref{fig:tracking_deg_reproj_ts}).
The corresponding \textit{factor} $\lambda$ is smaller than the threshold.
Therefore, we can swith the stable EM and then redo the tracking, improving the tracker's results.

The trajectories produced by the TS-based tracker on all sequences are compared 
in Fig. \ref{fig:simu_dof_plot}, Fig. \ref{fig:rpg_dof_plot}, and Fig. \ref{fig:upenn_dof_plot}.
Due to limited space, only the ground truth trajectories and trajectories of the tracker with the highest accuracy are shown as the reference.
Through these figures, readers can have a better understanding of the above explanations.
For example, we observe that the poses of the TS-based tracker "jump" on \textit{upenn\_indoor\_flying3} from Fig. \ref{fig:upenn_dof_plot}.
The unreliable TS degrades the performance.

    \subsection{Computation Time}
\label{sec:exp_computation_time}

The average computational time of tracker variations is reported in Table \ref{tab:latency}, 
where all of them take around $7$-$8ms$ to solve the pose estimation problem.
Depending on the sensor resolution, the creation of the TS takes about $5$-$10ms$ \cite{zhou2021event}.
The degeneracy evaluation takes about $0.5$-$1.5ms$.



    \subsection{Discussion}
\label{sec:exp_discussion}

All the tested trackers have their own limitations. 
First, although the enhanced TSEM-based tracker can mitigate the undesirable effect caused by the unreliable TS,
it suffers shortcommings of both the TS and EM if $\lambda_{th}$ is not well tuned.
This explains why the TSEM has bad accuracy on $\textit{simu\_poster\_6DoF}$, $\textit{simu\_checkerboard\_6DoF}$, and $\textit{rpg\_box}$.
Hence, the utilization of the TSEM is subtle: tunning a good parameter requies to know more about dataset, 
while the degeneracy check is used to evaluate the conditions in unfamiliar environments.

Second, we find that the estimated trajectories of the current tracker are not smooth. 
As an example, the TS-based tracker's trajectory on $\textit{rpg\_monitor}$ is visualized in Fig. \ref{fig:TS_traj_rpg_monitor}.
This is due to the fact that the tracker is sujected to large noise in optimization.
Introducing the motion prior \cite{forster2016svo}, considering historical measurements (sliding window-based framework \cite{qin2018vins}, 
full batch optimization \cite{mueggler2015continuous}, or regularization term \cite{liwicki2016coarse} could be possible
solution to solve this issue.
Moreover, as done in \cite{zihao2017event,rebecq2017real,vidal2018ultimate}, 
the aid of inertial measurements have significantly extended the working scope of monocular event camera.
Exploring stereo event-based VIO is also a promissing direction for this problem.

Finally, the image-type representations quantize event timestamps. The tracking results can be further optimized by introducing spline-based optimization framework \cite{mueggler2018continuous,gentil2020idol}.




\section{Conclusion}
\label{sec:conclusion}

In this work, we have conducted extensive comparisons of two image-type representations for event camera-based tracking.
We also introduce the degeneracy check and propose an enhanced tracker to make use of their complementary strengths.
Furthermore, we implement six tracker variations. 
In evaluating these trackers, our goal is to benchmark their performance in different-complexity scenarios.
The results presented in Section \ref{sec:experiment} suggest that each representation has its own advantages and limitations for tracking.
Such event-based representations are scene- and motion-dependent: both of them may degenerate in non-ideal situations.
We hope that these results and conclusions shown in this paper 
may provide a reference for researchers on event-based SLAM and indicate possible directions to improve the state-of-the-art (SOTA) methods.



{\small
\bibliographystyle{ieee_fullname}
\bibliography{egbib}
}

\end{document}